\documentclass[preprint,12pt,authoryear]{elsarticle}

\usepackage{amssymb}
\usepackage[normalem]{ulem}
\usepackage{babel}
\usepackage{amsmath}

\usepackage{graphicx}
\usepackage{fancyhdr}
\usepackage{multirow}
\usepackage{subcaption}
\usepackage{float}
\usepackage{url}
\usepackage{xcolor}
\usepackage{lineno}

\journal{Remote Sensing of Environment}

\begin{document}

\begin{frontmatter}



\title{Adaptive Modeling of Satellite-Derived Nighttime Lights Time-Series for Tracking Urban Change Processes Using Machine Learning}


\author[inst1]{Srija Chakraborty\corref{c1}}
\cortext[c1]{Corresponding Author: schakraborty@usra.edu}
\affiliation[inst1]{organization={Earth from Space Institute, Universities Space Research Association},
            addressline={425 3rd Street SW, Suite 950}, 
            city={Washington},
            postcode={20024}, 
            state={DC},
            country={USA}}
\author[inst1]{Eleanor C. Stokes}

\begin{abstract}
Remotely sensed nighttime lights (NTL) uniquely capture urban change processes that are important to human and ecological well-being, such as urbanization, socio-political conflicts and displacement, impacts from disaste\-rs, holidays, and changes in daily human patterns of movement. Though several NTL products are global in extent, intrinsic city-specific factors that affect lighting, such as development levels, and social, economic, and cultural charac\-teristics, are unique to each city, making the urban processes embedded in NTL signatures difficult to characterize, and limiting the scalability of urban change analyses. In this study, we propose a data-driven approach to detect urban changes from daily satellite-derived NTL data records that is adaptive across cities and effective at learning city-specific temporal patterns. The proposed method learns to forecast NTL signatures from past data records using neural networks and allows the use of large volumes of unlabeled data, eliminating annotation effort. Urban changes are detected based on deviations of observed NTL from model forecasts using an anomaly detection approach. Comparing model forecasts with observed NTL also allows identify\-ing the direction of change (positive or negative) and monitoring change severity for tracking recovery. In operationalizing the model, we consider ten urban areas from diverse geographic regions with dynamic NTL time-series and demonstrate the generalizability of the approach for detecting the change processes with different drivers and rates occurring within these urban areas based on NTL deviation. This scalable approach for monitoring changes from daily remote sensing observations efficiently utilizes large data volumes to support continuous monitoring and decision making. 
\end{abstract}




\end{frontmatter}


\section{Introduction}
\label{sec:sample1}
Nighttime lights (NTL) derived from remote sensing observations capture urban infrastructural dynamics signaling important human and environment\-al changes~(\cite{roman2018nasa, levin2020remote}). For example, NTL observations have been used to monitor infrastructural changes associated with urbanization~(\cite{xie2019temporal}), disasters impacting power grids~(\cite{wang2018monitoring, roman2019satellite, zhao2020time}), economic activity~(\cite{chen2011using,henderson2012measuring, nordhaus2015sharper}), socio-political conflicts and displacement~(\cite{bennett2017advances, li2013satellite, jiang2017ongoing}), holidays~(\cite{roman2015holidays}), human patterns of movement~(\cite{stokes2022tracking}), and impacts of unprecedented events such as COVID-19~(\cite{stokes2022tracking}). In particular, the Day/Night Band (DNB) onboard the Visible Infrared Imaging Radiometer Suite (VIIRS) on the Suomi-NPP and NOAA-20 platforms is uniquely suited to monitor both sudden and gradual urban changes. Sudden changes (e.g. disasters) are captured by its fine daily temporal resolution, and gradual changes (e.g. urbanization) are tracked due to the its decade long, and growing, imagery archive. However, changes in NTL over urban areas are particularly difficult to interpret. Multiple socio-economic and demographic factors introduce heterogeneity in NTL time-series across urban areas. Urban development level (impacting NTL magnitude and stability), cultural lighting practices (impacting NTL magnitude and distribution), and holidays and festivals (impacting seasonal patterns) render NTL signatures unique to a particular city. 
Dependence on such attributes also makes NTL time-series dynamic and continually evolving as urban areas change. Global scale monitoring of urban change, using NTL, is a challenge, as analyses must differentiate between usual and anomalous NTL patterns of change for each individual city. Dense remote sensing time-series of natural phenomena such as vegetation phenology, are often characterized using pre-defined models, such as harmonic functions, that describe the inter- and intra-annual periodicity of land cover reflectance~(\cite{jakubauskas2001harmonic, lhermitte2008hierarchical}). However, as NTL tracks infrastructure use, which is unique to each city, pre-defined functions are less suitable for describing the time-series in a scalable manner and detecting change. Consequently, scaling urban change detection requires approaches that can learn the usual NTL temporal signature (i.e. baseline variation) for each city in order to detect any deviations from this trajectory.

In this study, we propose a data-driven approach using neural networks to learn city-specific NTL time-series models of the expected baseline behavior that can be generalized across urban areas for monitoring urban infrastructur\-al change. These models learn to predict successive NTL sequences of a city in an unsupervised manner without any predefined function describing the dependence. We then use an anomaly detection approach that assumes that, in the absence of change, the models can successfully predict consecutive sequences that match the observed NTL. As NTL radiance diverges from baseline behavior during change, the actual observations deviate from the model predictions allowing change detection. Moreover, we also demonstrate that the manner in which the observations deviate from the model predictions can help characterize different change processes. We explore three neural network-based models and apply them to detect different urban infrast\-ructural change types, namely, power outage due to disasters, conflicts, and urbanization, across ten different urban areas. Our approach adapts to city-level variations making it scalable across multiple urban areas, does not require prior knowledge of change type, and successfully detects both positive/negative and gradual/abrupt changes. Moreover, by sequentially applying the trained models to incoming NTL time-series, these models also allow adjusting to temporal shifts in the trends of a city and monitoring using most recent observations unlike traditional data-driven approaches using decomposition. Monitoring time-series change using most recent observations can enable detecting multiple change components by highlighting additional deviations within an ongoing change segment (for e.g. power outage during a pandemic). Additionally, being an unsupervised approach, our method eliminates the need to obtain labeled datasets of change for training, which can be a major challenge in analyzing Earth Observation data~(\cite{small2021grand, persello2022deep}). As ground truth knowledge for many observed changes are unavailable, we retrospectively evaluate the detectors’ performance based on known change events over the corresponding time-steps. We have also operationalized this approach for detecting changes in NTL caused by the socio-economic impact of COVID-19 globally~(\cite{stokes2023unpub}). We discuss possible future extensions for enhancing the proposed approach and explore how it may enhance the characterization of a wider set of infrastru\-ctural changes. 

\subsection{Related Work}
Change detection is a widely studied area in remote sensing. A vast majority of the work is focused on bitemporal methods that analyze before af\-ter-change image pairs to detect change. While these methods are suitable for high spatial resolution observations that are sampled less frequently, the rich temporal information in dense satellite image time-series cannot be effectively extracted by these methods, nor can these methods characterize gradual changes, heterogenous, or incremental change well. Furthermore, for nighttime remote sensing, where day to day variation in radiance is high, bitemporal methods fail to differentiate between normal daily fluctuations in infrastructure use and significant change. 

Multitemporal analysis that can extract temporal information such as long-term trends, change dates, and recovery rates, are more suitable for monitoring dense time-series observations. Harmonic models that fit harmonic functions to describe temporal variations have been effective at characterizing the repetitive seasonal pattern of remote sensing time-series, such as from vegeta\-tion leaf on and leaf off, and land cover cycles~(\cite{jakubauskas2001harmonic, lhermitte2008hierarchical, kleynhans2009improving, anees2015relative, chakraborty2018time, zhu2020continuous}) and have recently been applied to NTL time-series~(\cite{li2022continuous, zheng2021characterizing, zheng2022impact}). Harmonic analysis provides a good fit to periodic patterns with relatively symmetric rises and falls, like most natural seasonal cycles, however there are several challenges in its application to urban NTL change detection. First, NTL fluctuates in a more complex manner, often having sharp peaks or drops and varies differently from time-series of natural physical variables~(\cite{zheng2021characterizing}). Second, the magnitude of NTL varies with urban size and development level, unlike reflectance and derived indices, which stay within a predefined range [0 to 1 or -1 to 1]. This makes it more challenging to model NTL time-series and adapt to drifts in incoming sequential data outside the baseline distribution.

In contrast, neural networks, have been shown to be effective at learning sequential patterns in a data-driven manner without prior assumption of the time-series evolution or shape. Convolutional neural networks have been applied to nighttime light images to 
automatically identify cloud-free observa\-tions over cities~(\cite{cao2022exploring}), but never to characterize urban infrastruc\-tural changes. 
Alternative data-driven approaches using Seasonal and trend decom\-position using Loess (STL) have been used to detect changes in NTL time series associated with holidays~(\cite{roman2015holidays}), power outag\-es after Hurricanes Irma and Maria in Puerto Rico~(\cite{zhao2020time}), urbanizati\-on in continental United States~(\cite{xie2019temporal}) and lockdowns in the Middle East during the COVID-19 pandemic~(\cite{ stokes2022tracking}). However, in each of these cases change is always detected with respect to a pre-determined “pre-event” baseline phase, which does not evolve over time. Neural networks differ from STL in that they learn the temporal dependence between successive windows and allow for predictions based on the most recent data, adapting to changes in the time-series. This enables them to detect anomalies, even in a continuou\-sly changing time-series. Here, we explore the applicability of multiple of neural network architectures to model dense nighttime light time-series from NASA’s Black Marble product~(\cite{roman2018nasa}) across multiple urban areas. We examine the approach over different change types and change rates and derive metrics that can be useful in tracking infrastructural processes across urban areas.  

\section{Methodology: Time-Series Anomaly Detection from Multi-step Ahead Neural Network Forecasts}

\begin{figure*}[h]
\centering
\resizebox{5in}{3in}{\includegraphics{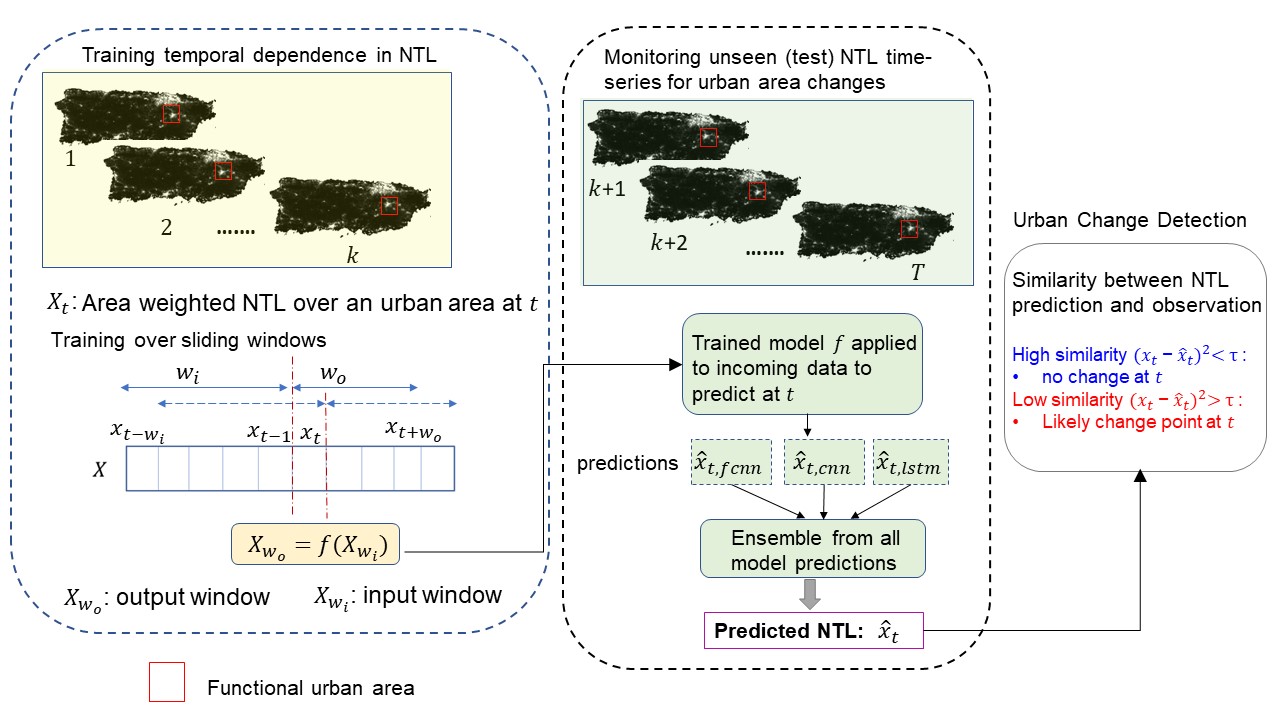}}
\vspace*{-.2mm}
\caption{Workflow of the proposed methodology to train the baseline temporal variation in NTL time-series between successive windows. The trained models are then applied to unseen test phase NTL observations where urban change processes are detected based on its deviation from model prediction. All steps of the approach are applied on an area weighted 1-D NTL time-series.}
\vspace*{-.5mm}
\label{fig:workflow}
\end{figure*}

Changes in nighttime lights from urban areas are marked by a deviation from the pre-change, expected NTL. 
To this end, we propose a data-driven approach that learns a model to forecast an NTL sequence (output window, $w_o$) given its preceding sequence (input window, $w_i$), creating a  multi ($w_o$) step ahead forecast. The model learns the temporal dependence between successive sequences, characterizing the expected NTL variation in a city in the absence of anomalies during the stable phase. The trained model is then applied to unseen incoming sequences to predict corresponding output sequences. Under no change conditions, the predicted output sequences are expected to closely match the true observed NTL resulting in a low prediction error. At time-steps of change, observed NTL deviates significantly from the prediction\-s, resulting in higher prediction error. By monitoring the prediction error we sequentially detect anomalous NTL observations by continually processing streaming NTL data. 
This approach enables learning city-specific models, that adapts to a city's socio-economic and cultural patterns captured through nighttime lights (NTL magnitude) and its temporal variability.

For each urban area considered in this study, the NTL time-series of length $T$ can be described as $X=\{x_1, x_2, \dots, x_T\}$, where $x_t$ represents the NTL radiance at time-step $t$. The objective of the proposed forecasting approach is to learn the temporal dependence between an input sequence $X_i=[x_{t-w_i},\dots, x_{t-1}]$ and its corresponding output sequence denoted as $X_o=[x_{t},\dots, x_{t+w_o}]$, using $X_o=f(X_i)$ and $w_o < w_i$ as shown in Figure~\ref{fig:workflow}. Here $f$ learns to capture the dependence between successive sequences from no-change phases in the training set.
The trained model is then applied to sequences of incoming NTL observations of the city.
Due to multi-step ahead forecasting, each time step $x_t$ is predicted $w_o$ times and the resulting prediction $\hat{x}_t$ is computed from the median of all $w_o$ predictions. Single-step head forecasting models were also used for predicting $X_o$, but were observed to be less suitable for detecting gradual changes and the approach was implemented using multi-step ahead models.


As the temporal dependence between input and output sequences is nonlin\-ear and city-specific, we propose to use neural network models which are effective at handling sequential data~(\cite{goodfellow2016deep},~\cite{bengio2009learning}) to learn $f$. 
Three different models are explored: fully connected neural networks (FCNN), 1-D convolutional neural networks (CNN) and long short-term memory neural networks (LSTM)~(\cite{hochreiter1997long}). Our approach is unsupervised, and each model is trained on naturally occurring variations in the baseline phase, leveraging large volumes of Black Marble time-series without requiring any labeling effort, that otherwise results in major bottlenecks.

\subsection{Ensemble NTL Prediction} 
After obtaining a forecast from the models,
a weighted average of each individual model's prediction is computed to derive the ensemble prediction. Ense\-mbles have been shown to enhance prediction accuracy, leading to more reliable forecasts and increased decision confidence~(\cite{dietterich2000ensemble},~\cite{ganaie2021ensemble}). We take a weighted ensemble where the most robust predictors (across no-change and change time-steps) 
are assigned the highest weights. The weights $c$ are held constant across all urban areas and are tuned 
by observing model performance. The derived ensemble prediction $\hat{x}_{t,ens}$ is expressed as as $\hat{x}_{t,ens}=\sum_{m} c_m \hat{x}_{t,m}$, where $m$ represents each model.


\begin{figure*}[h]
\centering
\resizebox{4.7in}{3in}{\includegraphics{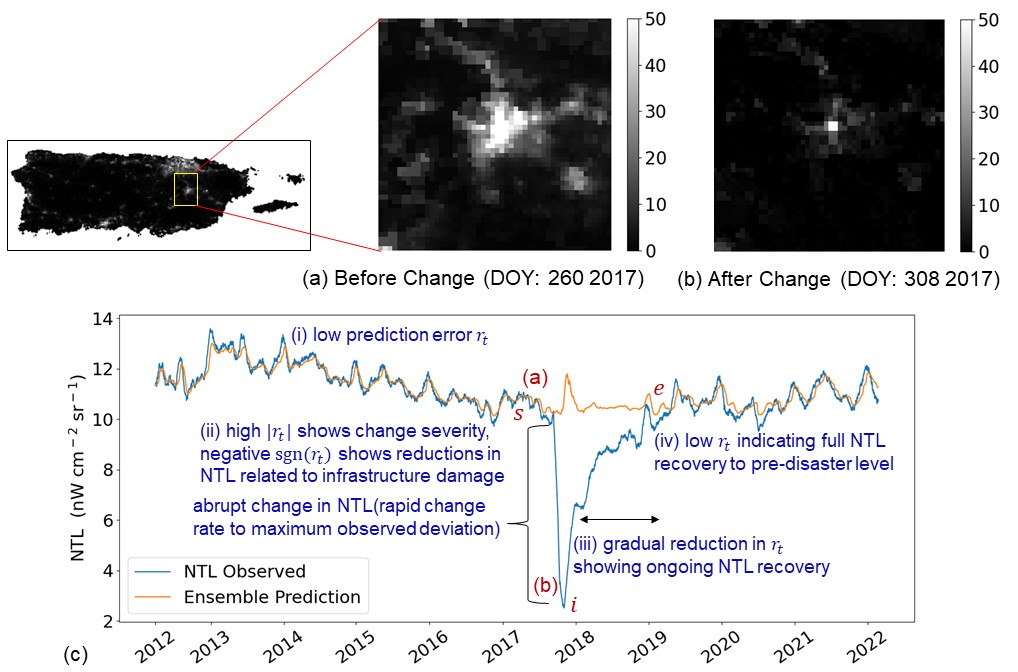}}
\vspace*{-.2mm}
\caption{Interpreting NTL time-series and derived metrics using the proposed approach. Example of (a) pre- and (b) post-disaster (Hurricane Maria) images of Caguas, Puerto Rico due to infrastructure damage resulting in power outage. (c) Rich temporal information captured through NTL time-series. By comparing the NTL observation at each time-step with the ensemble prediction, change metrics on severity, deviation, change rate and recovery inferences are extracted. Prediction error also shows different temporal stages in this urban area, namely (i) pre-change baseline, (ii) change, (iii) continuing recovery, (iv) full recovery to pre-change levels.}
\vspace*{-.5mm}
\label{fig:interpret}
\end{figure*}

\subsection{Urban Change Process Detection and Interpretation}
\label{sec:metrics}
Incoming NTL observations are monitored sequentially for change by comparing with model predictions. This difference is indicated by predict\-ion error and is given by
\begin{linenomath*}
\begin{equation}
    r_t=x_t-\hat{x}_{t,\text{ens}}.
    \label{eqn:ch}
\end{equation}
\end{linenomath*}
A change point in the NTL time-series is detected when the observed NTL shows a high deviation from the model forecast using $r_{t}^{2} > \tau$, where $\tau$ is a threshold that detects the top $\mathcal{T}\%$ mean squared prediction error over all time-steps for an urban area. For each change point we compute metrics or change indicators capturing rich temporal information describing urban processes as shown with an example in Figure~\ref{fig:interpret}. We measure change severity $|r_{t}|$, showing the magnitude or impact of change on NTL and the change direction $sgn(r_t)$ that relates to the nature of deviation observed in NTL at detected change points. A positive $sgn(r_t)$ shows an increase in NTL above the predictions from the ensemble, while a negative $sgn(r_t)$ is indicative of a reduction in NTL compared to the predictions. Monitoring change severity also allows tracking post-change behavior and any recovery or reversal to baseline levels. We also derive the change start rate $\lambda_s$ and end rates $\lambda_e$ given by 
\begin{linenomath*}
\begin{equation}
    \lambda_s=\frac{x_i-x_s}{i-s+1}, \text{and}
    \label{eqn:ch}
\end{equation}
\end{linenomath*}
\begin{linenomath*}
\begin{equation}
    \lambda_e=\frac{x_e-x_i}{e-i+1}.
    \label{eqn:ch}
\end{equation}
\end{linenomath*}
Here $s$, $e$ and $i$ represent the time-steps of change start, end and inflection point marking the time-step where change magnitude begins to reduce, respec\-tively. The rates indicate how abrupt or gradual the processes are. 
Additional\-ly, we also derive a decision confidence indicator based on the number of models that detected a change point.

\begin{table}[]
\centering
\caption{Study Areas of NTL Change Showing Change Type or Change Driver, Urban Areas Considered, Change Duration, Description and Observed Post-Change Behavior}
\label{ref:table_desc}
\resizebox{\linewidth}{!}{
\begin{tabular}{|l|l|l|l|l|l|}
\hline
                Change Type  & Urban Area & Known Duration &Observed Start& Description  &Post-Change Behavior  \\ \hline
\multirow{4}{*}{Disaster} & Beira, Mozambique  &03/2019-05/2019 & 03/2019&Negative, Abrupt & Full, rapid recovery  \\  \cline{2-6} 
                  & San Juan, Puerto Rico &09/2017-11/2018& 09/2017&Negative, Abrupt  & Full, gradual recovery  \\ \cline{2-6} 
                  & Ponce, Puerto Rico &09/2017-05/2018& 09/2017&  Negative, Abrupt & Full, gradual recovery  \\ \cline{2-6} 
                  & Caguas, Puerto Rico &09/2017-02/2019  & 09/2019&Negative, Abrupt  &Full, gradual recovery  \\ \hline
\multirow{3}{*}{Conflict} & Adwa, Ethiopia & 11/2020-02/2022& 11/2020 &Negative, Abrupt  & No recovery observed  \\ \cline{2-6} 
                  &Ad Dala, Yemen   & 09/2014-02/2022 & 03/2015 &Negative, Abrupt  & No distinct recovery\\ \cline{2-6} 
                  &Sana'a, Yemen  &09/2014-02/2022 & 04/2015 & Negative, Abrupt & No distinct recovery\\ \hline
\multirow{3}{*}{Urbanization} &Kathmandu, Nepal & 2020-2022 & 10/2019&Positive, Gradual  & Persistent increase  \\ \cline{2-6} 
                  &Arua, Uganda & 2020-2022 & 11/2019 &Positive, Gradual  & Persistent increase  \\ \cline{2-6} 
                  & Jinja, Uganda &2020-2022  & 12/2018 & Positive, Gradual & Fluctuating increase  \\ \hline
\end{tabular}}
\end{table}

\section{Experimental Details }
\subsection{Dataset}
We use the VNP46A2 dataset from NASA’s Black Marble Product Suite that records BRDF-corrected nighttime lights from the Day/Night Band onboard the VIIRS instrument~(\cite{roman2018nasa}). For each region of study, we access the full time-series record from January 19th 2012 up to 23rd February 2022 from NASA’s Level-1 and Atmosphere Archive and Distribution System Web Interface (LAADS-DAAC).
We then extract the gap-filled NTL pixel time-series from urban areas corresponding to each region of interest, which keeps the sample of pixels contributing to the mean urban radiance constant. Urban boundaries are defined using the Global Human Settlements Functional Urban Areas dataset (GHS-FUA)~(\cite{moreno2021metropolitan}). For each urban polygon, we derive a daily net area weighted average of NTL in units of $\text{nW} \; \text{cm}^2 \; \text{sr}^{-1}$. The area of each pixel is obtained using the WGS84 datum. This daily NTL time-series is smoothed using a 30-day rolling average. 


\subsection{Change Areas and Training Sets}

 We considered three different categories of urban changes captured by NTL, namely, (1) the impact of disasters on urban infrastructure, (2) conflicts, and (3) urbanization. The urban areas considered for each change type are shown in Table 1. To evaluate the method, for each case the “ground-truth” change date is estimated from news stories as the first day of an event (e.g. the first battle in a war for conflict or the day a hurricane made landfall). Urbanization, being a long term change, has no exact start or end date, so for accuracy assessment purposes we estimate the year that population began trending upward from available census for Kathmandu, Nepal~(\cite{PopStat})~(\url{https://populationstat.com/nepal/kathmandu}). For and Arua and Jin\-ja, Uganda, we accessed records from~(\cite{CityPop})~(\url{https://www.citypopulation.de/en/uganda/cities/}) and allow a $\pm$ 1 year buffer when evaluating NTL change results caused by urbanization.

As training over pre-change observations is essential for the models to learn the baseline variation, we retrospectively set the training end dates such that they lie sufficiently before the ground-truth change time step. For each urban area, we include all observations from the start of the time-series in 2012 to the retrospectively determined training end date, which at minimum consists of three years of daily NTL data. In all cases, our test phase consists of both `no-change’ and `change’ states allowing us to evaluate the model behavior in both states.

\subsection{Model Training}
The models are trained on an input window $w_i$ of 60 days of NTL observat\-ions and learn to predict NTL sequences with an output window $w_o$ of length 30. The input and output sequence pairs from the training set representing the pre-change baseline are further split with 80\% of the data used for training and the remaining 20\% used for validation. 
We use the TensorFlow, Keras environment and train each model to predict successive sequences using the Adam optimizer~(\cite{kingma2014adam}) and the mean absolute error loss. We used graphical processing units (NVIDIA-GeForce-RTX-3090) to accelerate training. For change detection, we set $\mathcal{T}=25$ to select the top 25\% prediction error so that both abrupt and gradual processes can be detected. Further experimental details are described in~\ref{sec:appendix}.

\section{Evaluation}
\label{ref:sec_eval}
We quantify the performance of the change detectors with commonly used metrics for evaluating detection. Change time-steps are considered the `positive’ class while non-change time-steps are considered `negative’. For each change event date, the known change duration is considered as the ground truth for the positive class. We compute the true positive rate or recall as 
\begin{linenomath*}
\begin{equation}
    recall (R)=\frac{TP}{TP+FN},
    \label{eqn:ch}
\end{equation}
\end{linenomath*}
where TP indicates true positives, and FN, False Negatives. We also measure detection delay $\delta$ defined as the number of time-steps from the onset of change it takes for the detectors to identify a change point. 

NTL time-series consist of several time-steps with a high (positive or negative) signal deviation that cannot be attributed to a known event. In particular, ground truthing the true negative class requires extensive local knowledge of the urban area and accurate historical records of not only events that happened, but of normalcy in lighting patterns. As such, we define “no change” time-steps as those where the NTL signal does not deviate more than 10\% from the city’s median NTL baseline and mark a detection as a false positive when these time-steps are identified as change points. 
We then compute precision using
\begin{linenomath*}
\begin{equation}
    precision (P)=\frac{TP}{TP+FP},
    \label{eqn:ch}
\end{equation}
\end{linenomath*}
where FP denotes false positives. As the fraction of change time-steps is much lower than non-change time-steps, class imbalance has to be taken into account to evaluate the models' effectiveness at detecting change~(\cite{provost2001robust}). Recall and delay are more suitable to evaluate the detectors for correctly identifying change as they relate to the performance with the positive, change class. We also derive the F${_\beta}$ measure that derives the weighted harmonic mean of precision and recall with $\beta$=2 to assign higher cost for undetected change time-steps through higher weight on recall. We report all metrics for disasters and conflicts in daily units, while urbanization is reported in yearly units.

\section{Results}
\begin{table}[t]
\centering
\caption{Evaluating Detection Performance Using Recall, Precision, F-$\beta$, Delay}\label{ref:table-eval}
\renewrobustcmd{\bfseries}{\fontseries{b}\selectfont}
\renewrobustcmd{\boldmath}{}
\scalebox{0.565}{
\begin{tabular}{|l|llll|llll|llll|llll|}
\hline
\multirow{2}{*}{Site} & \multicolumn{4}{l|}{Ensemble}                            & \multicolumn{4}{l|}{FCNN}                            & \multicolumn{4}{l|}{CNN}                            & \multicolumn{4}{l|}{LSTM}                            \\ \cline{2-17} 
                  & \multicolumn{1}{l|}{$R$(\%)} & \multicolumn{1}{l|}{$P$(\%)} &\multicolumn{1}{l|}{F$_\beta$(\%)} &$\delta$  & \multicolumn{1}{l|}{$R$(\%)} & \multicolumn{1}{l|}{$P$(\%)} &\multicolumn{1}{l|}{F$_\beta$(\%)}& $\delta$ & \multicolumn{1}{l|}{$R$(\%)} & \multicolumn{1}{l|}{$P$(\%)} &\multicolumn{1}{l|}{F$_\beta$(\%)} &$\delta$ & \multicolumn{1}{l|}{$R$(\%)} & \multicolumn{1}{l|}{$P$(\%)}&\multicolumn{1}{l|}{F$_\beta$(\%)} & $\delta$ \\ \hline
                  Beira & \multicolumn{1}{l|}{ 100} & \multicolumn{1}{l|}{ 13.18}  &\multicolumn{1}{l|}{ 43.15}& 0  & \multicolumn{1}{l|}{97.14} & \multicolumn{1}{l|}{13.14}&\multicolumn{1}{l|}{42.63}&2  & \multicolumn{1}{l|}{100} & \multicolumn{1}{l|}{12.43}&\multicolumn{1}{l|}{41.51} & 0 & \multicolumn{1}{l|}{87.54} & \multicolumn{1}{l|}{11.53}&\multicolumn{1}{l|}{37.76} &2  \\ \hline
                  San Juan& \multicolumn{1}{l|}{100} & \multicolumn{1}{l|}{84.50}&\multicolumn{1}{l|}{96.46} & 0 & \multicolumn{1}{l|}{100} & \multicolumn{1}{l|}{87.03}&\multicolumn{1}{l|}{97.10} &  0 & \multicolumn{1}{l|}{100} & \multicolumn{1}{l|}{ 90.83}&\multicolumn{1}{l|}{98.02} &  0 & \multicolumn{1}{l|}{100} & \multicolumn{1}{l|}{72.19}&\multicolumn{1}{l|}{92.98} & 0 \\ \hline
                  Ponce& \multicolumn{1}{l|}{99.58} & \multicolumn{1}{l|}{98.74}&\multicolumn{1}{l|}{ 99.74} & 1 & \multicolumn{1}{l|}{99.57} & \multicolumn{1}{l|}{98.74}&\multicolumn{1}{l|}{99.40} & 1 & \multicolumn{1}{l|}{99.16} & \multicolumn{1}{l|}{ 99.16}&\multicolumn{1}{l|}{ 99.16} &2  & \multicolumn{1}{l|}{97.89} & \multicolumn{1}{l|}{90.63}&\multicolumn{1}{l|}{96.34} & 1 \\ \hline
                  Caguas& \multicolumn{1}{l|}{ 100} & \multicolumn{1}{l|}{70.53}&\multicolumn{1}{l|}{92.29} & 0 & \multicolumn{1}{l|}{98.10} & \multicolumn{1}{l|}{67.98}&\multicolumn{1}{l|}{89.83} & 0 & \multicolumn{1}{l|}{ 100} & \multicolumn{1}{l|}{77.45}&\multicolumn{1}{l|}{ 94.5} &  0 & \multicolumn{1}{l|}{99.57} & \multicolumn{1}{l|}{68.6}&\multicolumn{1}{l|}{91.32} & 1  \\ \hline
            Adwa      & \multicolumn{1}{l|}{100} & \multicolumn{1}{l|}{90.65} &\multicolumn{1}{l|}{97.98}& 0 & \multicolumn{1}{l|}{87.16} & \multicolumn{1}{l|}{92.00}&\multicolumn{1}{l|}{88.08} & 4 & \multicolumn{1}{l|}{100} & \multicolumn{1}{l|}{91.7} &\multicolumn{1}{l|}{98.22}& 0 & \multicolumn{1}{l|}{100} & \multicolumn{1}{l|}{87.48}&\multicolumn{1}{l|}{97.22} & 0 \\ \hline
            Ad Dala& \multicolumn{1}{l|}{33.27} & \multicolumn{1}{l|}{100}&\multicolumn{1}{l|}{38.39} & 202 & \multicolumn{1}{l|}{33.27} & \multicolumn{1}{l|}{100}&\multicolumn{1}{l|}{38.39} & 224 & \multicolumn{1}{l|}{33.27} & \multicolumn{1}{l|}{100}&\multicolumn{1}{l|}{38.39} & 222 & \multicolumn{1}{l|}{32.02} & \multicolumn{1}{l|}{99.31}&\multicolumn{1}{l|}{37.04} & 191 \\ \hline
            Sana'a& \multicolumn{1}{l|}{33.27} & \multicolumn{1}{l|}{100} &\multicolumn{1}{l|}{38.39}& 257 & \multicolumn{1}{l|}{33.27} & \multicolumn{1}{l|}{100} &\multicolumn{1}{l|}{38.39} & 274 & \multicolumn{1}{l|}{33.27} & \multicolumn{1}{l|}{100}&\multicolumn{1}{l|}{38.39} & 215 & \multicolumn{1}{l|}{33.27} & \multicolumn{1}{l|}{100}&\multicolumn{1}{l|}{100} & 223 \\ \hline
                  Kathmandu& \multicolumn{1}{l|}{100} & \multicolumn{1}{l|}{100}&\multicolumn{1}{l|}{100} & 0 & \multicolumn{1}{l|}{100} & \multicolumn{1}{l|}{100}&\multicolumn{1}{l|}{100} & 0 & \multicolumn{1}{l|}{100} & \multicolumn{1}{l|}{100}&\multicolumn{1}{l|}{100} & 0 & \multicolumn{1}{l|}{100} & \multicolumn{1}{l|}{100}&\multicolumn{1}{l|}{$p$(\%)} & 0 \\ \hline
                  Arua& \multicolumn{1}{l|}{100} & \multicolumn{1}{l|}{100}&\multicolumn{1}{l|}{100} & -1 & \multicolumn{1}{l|}{100} & \multicolumn{1}{l|}{100}&\multicolumn{1}{l|}{100} & -1 & \multicolumn{1}{l|}{100} & \multicolumn{1}{l|}{100}&\multicolumn{1}{l|}{100} & -1 & \multicolumn{1}{l|}{100} & \multicolumn{1}{l|}{100}&\multicolumn{1}{l|}{100} &-1  \\ \hline
                  Jinja& \multicolumn{1}{l|}{100} & \multicolumn{1}{l|}{100}&\multicolumn{1}{l|}{100} & -1 & \multicolumn{1}{l|}{100} & \multicolumn{1}{l|}{100}&\multicolumn{1}{l|}{100} &-1  & \multicolumn{1}{l|}{100} & \multicolumn{1}{l|}{100}&\multicolumn{1}{l|}{100} & -1 & \multicolumn{1}{l|}{100} & \multicolumn{1}{l|}{100}&\multicolumn{1}{l|}{100} &-1  \\ \hline
\end{tabular}}
\footnotetext{test}
\end{table}

\begin{figure*}[ht]
\begin{subfigure}{.5\textwidth}
  \centering
  \includegraphics[width=0.8\linewidth]{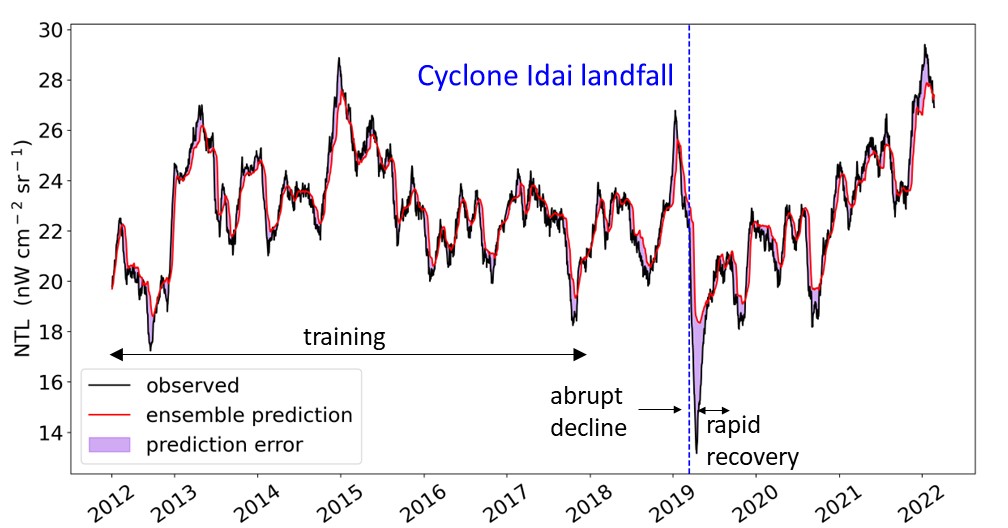}
  \caption{Cyclone Idai in Beira}
  \label{fig:sfig1}
\end{subfigure}%
\begin{subfigure}{.5\textwidth}
  \centering
  \includegraphics[width=0.8\linewidth]{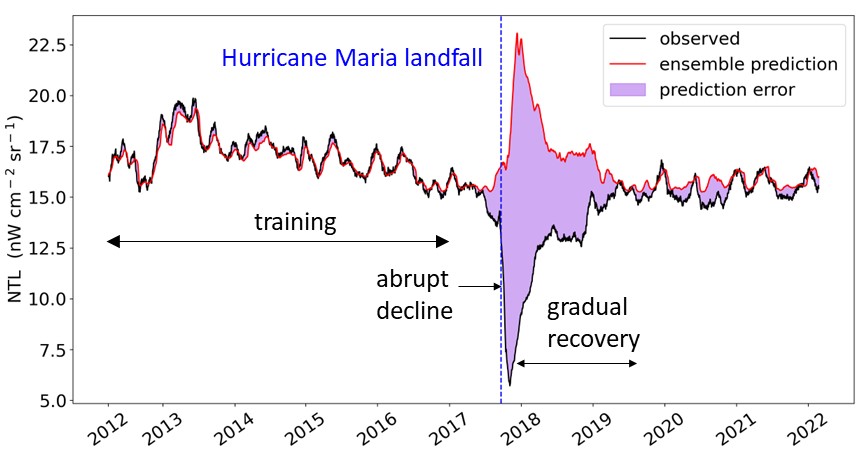}
  \caption{Hurricane Maria in San Juan}
  \label{fig:sfig2}
\end{subfigure}
\begin{subfigure}{.5\textwidth}
  \centering
  \includegraphics[width=0.8\linewidth]{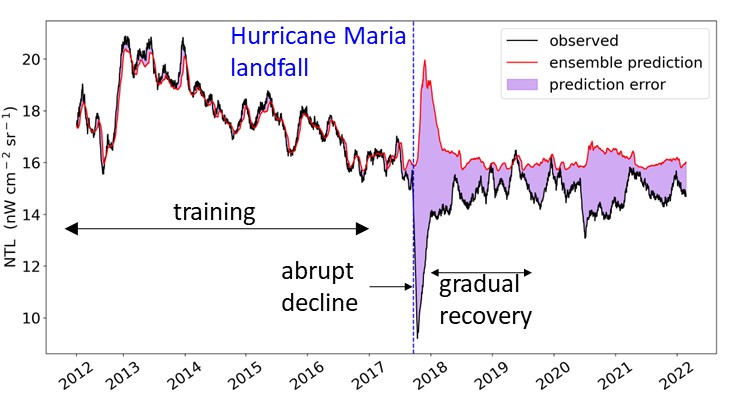}
  \caption{Hurricane Maria in Ponce}
  \label{fig:sfig3}
\end{subfigure}
\begin{subfigure}{.5\textwidth}
  \centering
  \includegraphics[width=0.8\linewidth]{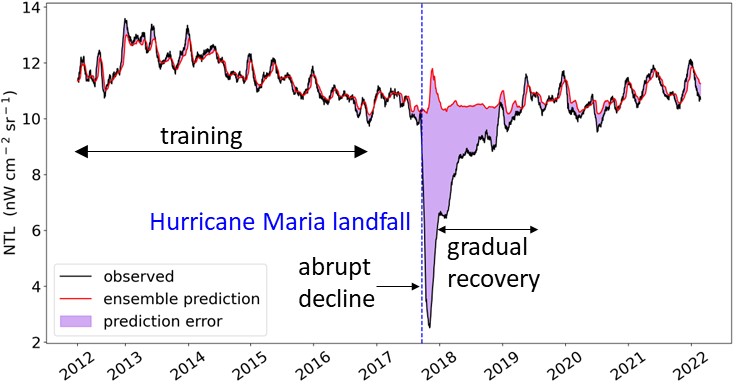}
  \caption{Hurricane Maria in Caguas}
  \label{fig:sfig4}
\end{subfigure}
\caption{Power outages from natural hazards as seen through NTL time-series in the four study areas. This change type produces abrupt drop with high change severity (as shown in the shaded area) followed by recovery.}
\label{fig:fig_dis}
\end{figure*}

\begin{figure*}[ht]
\begin{subfigure}{.31\textwidth}
  \centering
  \includegraphics[width=1.01\linewidth]{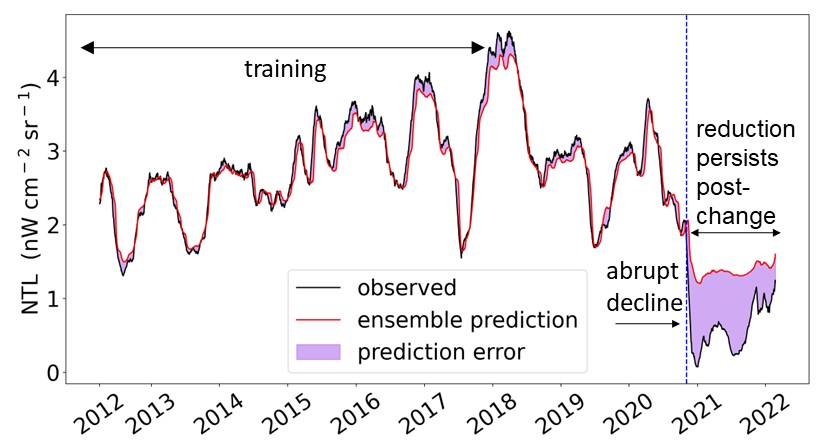}
  \caption{Conflict in Adwa}
  \label{fig:sfig5}
\end{subfigure}%
\begin{subfigure}{.32\textwidth}
  \centering
  \includegraphics[width=1.01\linewidth]{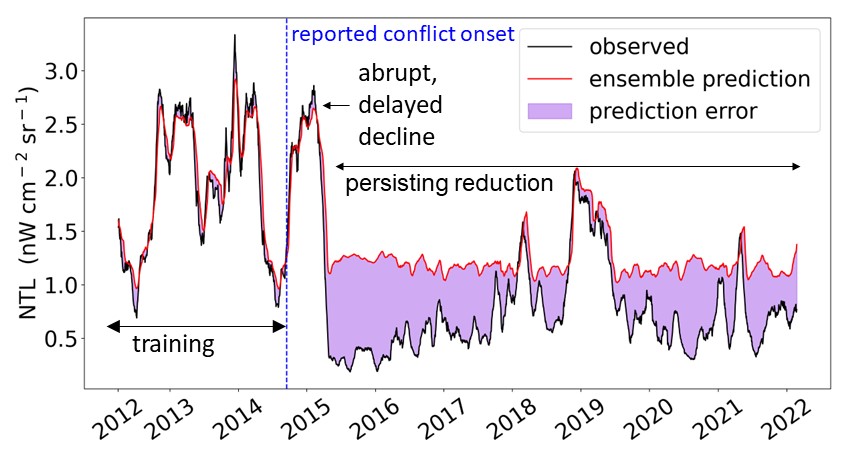}
  \caption{Conflict in Ad Dala}
  \label{fig:sfig6}
\end{subfigure}
\begin{subfigure}{.31\textwidth}
  \centering
  \includegraphics[width=1.03\linewidth]{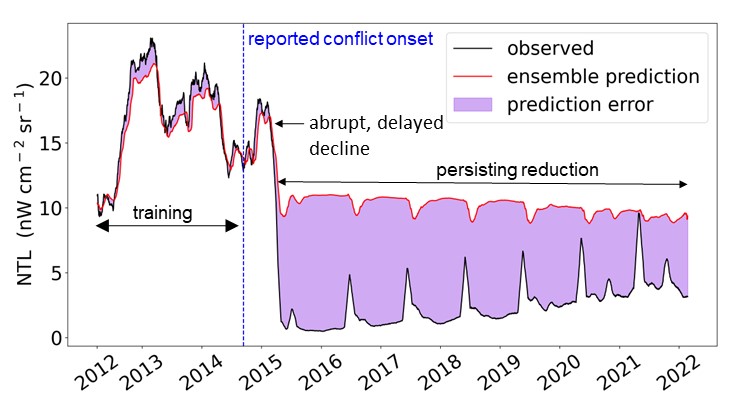}
  \caption{Conflict in Sana'a}
  \label{fig:sfig7}
\end{subfigure}
\caption{Impact of socio-economic changes caused by conflict as seen through NTL. Conflict is characterized by sudden negative and sustained deviation (as shown in the shaded area). Negligible recovery trends are observed and the NTL time-series stabilizes at the post-conflict levels.}
\label{fig:fig_conf}
\end{figure*}

\begin{figure*}[ht]
\begin{subfigure}{.31\textwidth}
  \centering
  \includegraphics[width=1.02\linewidth]{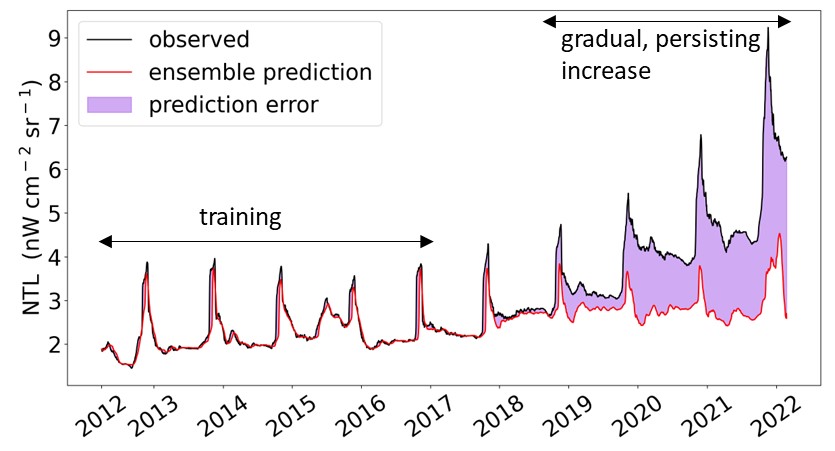}
  \caption{Urbanization in Kathmandu}
  \label{fig:sfig8}
\end{subfigure}%
\begin{subfigure}{.32\textwidth}
  \centering
  \includegraphics[width=1.02\linewidth]{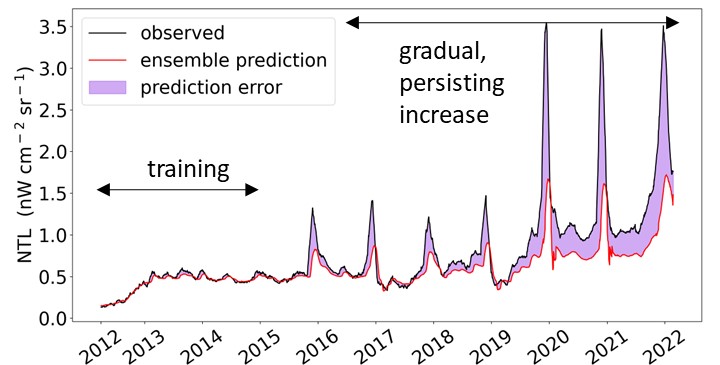}
  \caption{Urbanization in Arua}
  \label{fig:sfig9}
\end{subfigure}
\begin{subfigure}{.32\textwidth}
  \centering
  \includegraphics[width=1.02\linewidth]{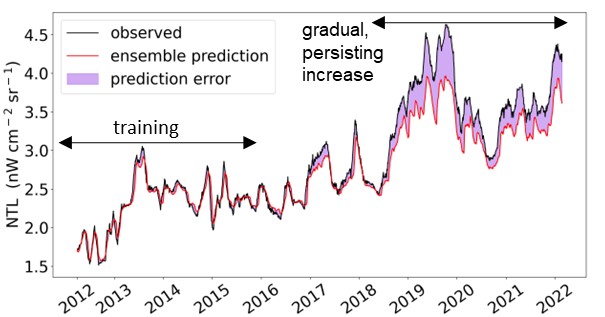}
  \caption{Urbanization in Jinja}
  \label{fig:sfig10}
\end{subfigure}
\caption{Changes caused by urbanization as seen through NTL time-series. Urbanization is characterized by gradual, positive deviation (as shown in the shaded area). A continuing increase in NTL is seen in the post-change onset stage.}
\label{fig:fig_urb}
\end{figure*}

The change detection individual model and ensemble performance are summarized in Table~\ref{ref:table-eval} using the metrics from Section~\ref{ref:sec_eval}. We observe a high recall (R) (average of 86.61\% across all areas) indicating that the ensemble correctly identifies change. We note a high precision (P) (average of 85.76\% across all areas) showing low false positives, indicating the robustness of the detectors. The F$_\beta$ measure with $\beta=2$ shows the combined detection performance with higher weight on recall to assign larger penalty for missing change points. We observe comparable performan\-ce by all three models with an average standard deviation of 1.35\% in recall. However, we observed that the stability of the predictions is the highest with LSTM, while CNN predictions exhibited higher instability particularly at change points. Therefo\-re, we assigned $c_{\text{LSTM}}=0.5$, $c_{\text{ANN}}=0.3$ and $c_{\text{LSTM}}=0.2$ as ensemble weights.

Figures~\ref{fig:fig_dis}-\ref{fig:fig_urb} show the comparison of observed NTL and predicted NTL for each urban area, that shows the variation in prediction error. For a given type of change, the shape of the prediction error time-series is similar for across different urban areas and thus can be used for change type characterization. 

\begin{figure*}[h]
\centering
\resizebox{4.0in}{3.5in}
{\includegraphics{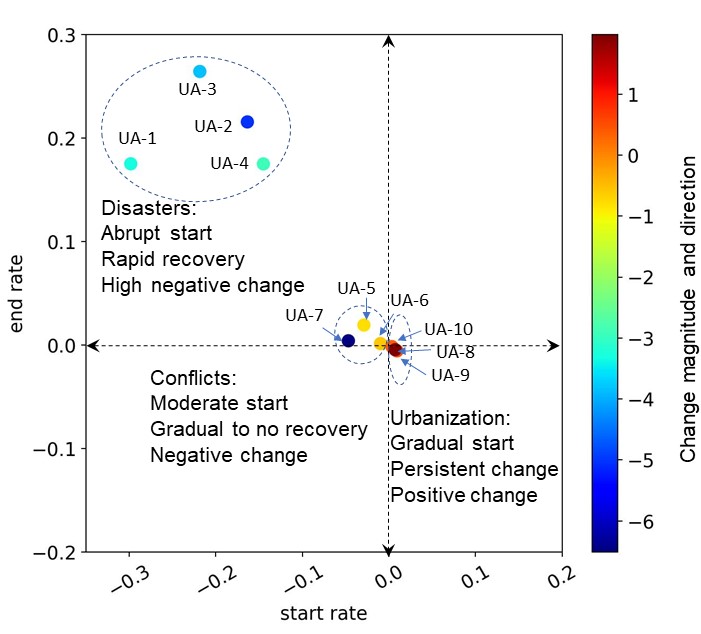}}
\vspace*{-.2mm}
\caption{Comparing change types using metrics extracted from prediction error time-series. A plot of change start rate against change end rate contrasted with the average magnitude and direction of change shows salient properties of each change type. Here magnitude of change shows the average prediction error during change representative of severity while direction shows the deviation direction compared to ensemble predictions.}
\vspace*{-.5mm}
\label{fig:change_signature}
\end{figure*}

Disasters cause an abrupt decline in NTL due to power outages resulting in high prediction error $|r_t|$, at the onset of change as seen in {Figures}~\ref{fig:sfig1},~\ref{fig:sfig2},~\ref{fig:sfig3}, \ref{fig:sfig4}. The change direction indicated by $sgn(r_t)$ is negative due to disaster-induced infrastructure damage. As $r$ is largest at the onset, disasters are detected by the models with minimum or no delay and align with the known change dates. 
For Beria City, Mozambique in Figure~\ref{fig:sfig1}, the post-change NTL shows rapid recovery to pre-change levels, while NTL in San Juan, Puerto Rico in Figure~\ref{fig:sfig2}, Ponce, Puerto Rico in Figure~\ref{fig:sfig3}, and Caguas, Puerto Rico in Figure~\ref{fig:sfig4} show relatively slower recovery. High precision, recall and F-$\beta$ scores and low detection delay are noted in Table~\ref{ref:table-eval}. A low precision in Beira, Mozambique can be attributed to a relatively unstable baseline time-series, resulting in an increase in false positives. However, several false positive detections are transient disturbances and can be identified as contaminations by noting the short temporal persistence of these occurrences.

Conflict causes a steep and long-lasting reduction in NTL (Figures~\ref{fig:sfig5}, \ref{fig:sfig6}, \ref{fig:sfig7}) as shown in the large deviation in prediction error $|r_t|$ during change. 
However, the reduction in NTL can be delayed from the known onset of conflict due to the time it takes to accrue infrastructure damage  (e.g. $\delta =6$ months in Ad Dala, Yemen in Figure~\ref{fig:sfig6}; $\delta=7$ months in Sana’a, Yemen Figure~\ref{fig:sfig7}). In addition, the delay can be caused by inaccurate or incomplete knowledge of the spatio-temporal progression of conflict, causing errors in the ground-truth. For the conflict in Adwa, Ethiopia, the detected change date aligns with the known conflict date. This can be explained by the relatively small size of the Tigray region ($\approx$ 19,000 square miles), where the conflict is occurring. In comparison, the known conflict start date over the area of Yemen ($\approx$ 530,000 square miles) does not account for the sub-national spread among cities, producing a delay in observed decline in NTL. Satellite-based monitoring can thus provide insights about the spread of conflicts across different cities, especially in remote regions, improving the spatial resolution of ground-based records. Conflict is characterized by an abrupt reduction in NTL with a moderately high change start rate. As with disasters, the change direction $sgn(r_t)$ is negative but the rate of recovery is slower than for outages caused by disasters. In Ad Dala (Figure~\ref{fig:sfig6}), no consistent recovery is observed. In Sana’a (Figure~\ref{fig:sfig7}), short-lived seasonal peaks centered around Ramadan are observed, and a very gradual recovery trend from 2019 which has not yet returned to pre-change NTL levels. For Adwa (Figure~\ref{fig:sfig5}), a very gradual recovery signal is observed, but cannot be confirmed as the post-change observation period is short. We observe high precision, recall, F-$\beta$ and low delay for Adwa, Ethiopia as shown in Table~\ref{ref:table-eval}, while low recall, high delay are observed in Ad Dala, Yemen and Sana’a, Yemen due to the incomplete ground truth record of sub-national spread of conflict. 

Urbanization results in a gradual increase in NTL as seen from $r_t$, with the increasing trend lasting for years producing a positive deviation, between observed NTL and model forecasts. Predictions in Kathmandu, Nepal  shows a continually increasing NTL signal~(Figure~\ref{fig:sfig8}) that agrees with the yearly population reports. Change detection begins in October 2019, aligning closely with the ground-truth. 2020 census records for Kathmandu show a 36\% increase in population compared to 2012, a trend that continues through the most recent NTL data in 2022.
Urbanization is also detected in Arua, Uganda starting in August, 2019~(Figure \ref{fig:sfig9}), and in Jinja, Uganda starting in December 2018~(Figure \ref{fig:sfig10}). Census data for Uganda is not available at the urban level on a yearly basis, however Arua and Jinja had a 2.7\% and 1.6\%  annual population growth rate between the years of 2014 and 2020, respectively. NTL tracks the growth of electrified infrastructure that accompanies urban population growth and development, as has been noted in previous studies~(\cite{zhou2015global}). In all cases, urbanization is observed to be a gradual change process with relatively lower magnitude of $|r_t|$ compared to the other change types. High precision, recall, F-$\beta$ and low delay is noted for all urbanization cases. For Arua, Uganda and Jinja, Uganda, we detect the urbanization signal a year before (2019) ($\delta$=-1 in Table~\ref{ref:table-eval}) the available population records (2020).

Although high quality ground truth data was not available for all events, we still observed an average decision agreement of about 86.34\% over the study duration between the models’ change detection date and the known or approximated change dates serving as ground truth. 
Overall, this shows the effectiveness of the methods at adjusting to a city’s baseline, adapting to the variation that is embedded in temporal windows and sequentially tracking deviations caused by change. By using a one-class anomaly detection approach, the models successfully monitor the NTL time-series without any prior knowledge of change. Moreover, the average prediction error, its directi\-on and change rates are consistent for each change type~(Figure~\ref{fig:change_signature}) and can thus be viewed as characteristic temporal signatures of each change type. This can be useful in classifying and interpreting different change types after detection.

~\section{Discussion}
This study introduced a novel, adaptive, data-driven forecasting-based anomaly detection approach to detect and characterize urban change process\-es from NTL time-series. 
The approach is applicable across cities and adapts to city-specific characteristics embedded in NTL. 
By forecasting successive NTL sequences using a moving window, the approach also allows continually adjusting to trends in the time-series to effectively monitor the streaming NTL data records. The high decision agreement of 86.34\% between the models with different architectures show that data-driven approaches effectiv\-ely learn the embedded temporal patterns and the decisions are relatively invariant to model selection. 
As the models do not require knowledge of change signature, they are extendable to other change classes that are captur\-ed by NTL such as global socio-economic response due to events such as COVID-19~(\cite{stokes2023unpub}) as well as electrification trends across urban areas. 
The use of a long-term time-series also allows continual monitori\-ng of change severity, change direction and post-change behavior. The approach proposed here can provide interpretable change metrics for a wide range of downstream applications such as informing and monitoring recovery from disasters, tracking socio-economic changes and mapping vulner\-able locations as a result of conflicts required for planning humanitarian efforts~(\cite{blind2019humanitarian}) on a near-real time scale to support decision making. In addition, it can be applied to monitor gradual processes such as urbanization and electrification.  

One advantage of neural network-based forecasting is its ability to extract complex city-level signatures and adapt to the variations of an urban area for scaling geographically. This is enabled through the training step where the models learn the temporal attributes of a given urban area from large volumes of past data by utilizing advanced computational platforms. We utilized graphical processing units in a distributed computing environment for training the models and observed our proposed approach to be computatio\-nally scalable, requiring an average of $50$ seconds for training and predicting NTL sequences and $5$ seconds for deriving per time-step change metrics for each urban area. 
By leveraging advanced compute facilities, the proposed methods are generalizable and scalable for global analysis as shown in~(\cite{stokes2023unpub}). Moreover, our approach eliminates expensive labeling steps which can introduce bottlenecks in processing remote sensing datasets.

While model agreement is indicated through ensemble decision confidence to indicate the reliability of the predictions, any derived estimates depend on the input data quality. We have utilized the highest quality, stable, daily NTL product available. However, successive periods of poor quality retrievals caused by prolonged periods of cloud cover can affect retrievals, trained models and predictions. Given that there are known errors in the cloud mask, and variation in daily NTL introduced from sensor angle~(\cite{wang2021quantifying}), we expect the detection to be improved with the release of the Black Marble collection 2 product.  
As our monitoring is based on an anomaly detection approach, the assumption is that urban change processes introduce deviations in the data that are larger than baseline period variation. While the anomaly detectors are robust to transient noise and aberrations, lasting deviations, particularly if unexplained, are expected to impact model training. Improved retrievals and NTL data quality, can also enhance the robustness of the detectors. Urban change processes that cause deviations that are smaller than the baseline variations may be undetected. 
In scenarios where the NTL baseline shifts considerably after a change, retraining using concept drift strategies will be required so the models can adjust to the distinct post-change NTL distribution.

\section{Conclusion}
This study introduces a novel data-driven approach to identify urban change from satellite-derived estimates of urban process change using nightti\-me light time-series. We utilize advances in machine learning and learn multiple neural network-based models that adapt to a city’s NTL signature representing a baseline model defining variations between successive NTL sequences. These models are then used to derive a prediction for sequences of incoming observations and to form an ensemble forecast that is monitored sequentially for deviations from the observed NTL to detect change. We demonstrate this approach to be effective at detecting both gradual and abrupt urban changes in an unsupervised manner from different geographic regions, and for deriving metrics that can aid in classifying and interpreting change processes.

While we explored three classes of neural networks for deriving the ensemb\-les, additional variants for handling sequential data can also be adapted into our workflow to further improve the robustness of our models. By training on the NTL sequences of detected change and using techniques such as metric learning, we will explore strategies to minimize the training requirement for generalizing the models in the future. The derived interpretations of change described in Section~\ref{sec:metrics} and shown in Figure~\ref{fig:change_signature} act as temporal signatures of urban change and can be used along with prediction error time-series shape analysis for classifying different change types. Our approach will also be extended to pixel-level NTL time-series analysis for automated spatio-temporal monitoring of intra-urban infrastructural changes. Furthermore, by integrating with higher spatial resolution datasets that usually have limited temporal resolution, the proposed multitemporal analysis can be utilized to infer the context of change. Complementary information from higher resolution image\-ry, synthetic aperture radar can further help in classifying change types. Overall, by analyzing dense NTL time-series, the proposed approach can extract long term trends to monitor urban areas continually and assist domain scientists and policymakers in decision making by monitoring large volumes of satellites data. 


\section*{Acknowledgements}
Funding for this project was provided by NASA's Rapid Response and Novel Research in Earth Science (RRNES) grant 80NSSC20K1083. Computa\-tional resources on the National Research Platform were obtained from the Research Institute for Advanced Computer Science, USRA.

\bibliographystyle{elsarticle-harv} 
\bibliography{elsarticle-template-harv}

\appendix
\section{Experimental Details and Forecasting Models}
\label{sec:appendix}

To train the baseline models using neural networks, we use an Adam optimizer and mean absolute error loss to penalize when the predictions fail to emulate the observation behavior in the baseline phase for input output pair in the training set. We use regularization techniques (dropout, batch normalization, activity and kernel constraints and activity regularizers) to reduce overfitting. The models (FCNN, CNN, and LSTM) are trained over 70, 90 and 25 epochs respectively and set based on approximate plateauing of the validation loss. The model hyperparameters are determined experimentally from the performance over the validation set such that these can adapt across urban areas.

The neural network architectures can be described as follows: 

CNN: This architecture consists of four 1-D convolut\-ional layers with a maxpooling, batch normalization and 10\% dropout after each layer. The convolutional layers consist of 90, 45, 30 20 filters with a kernel size of 9, 9, 6, 6 respectively. The convolutional stacks are flattened and followed by three dense layers with 20, 15 and 30 units. 

FCNN: This architecture consists of four dense layers with 60, 45, 25, 30 units, respectively followed by 10\% dropout after each layer. 

LSTM: The LSTM architecture consists of two LSTM layers with 45 and 30 units followed by 10\% dropout after each and are connected with three dense layers with 30, 15 and 30 units that produce the predictions.

All models use the ReLU activation and a batch size of 64.





\end{document}